\documentclass{article}
\usepackage{ijcai23}
\usepackage{times}
\usepackage{soul}
\usepackage{url}
\usepackage[hidelinks]{hyperref}
\usepackage[utf8]{inputenc}
\usepackage[small]{caption}
\usepackage{graphicx}
\usepackage{amsmath}
\usepackage{amsthm}
\usepackage{booktabs}
\usepackage{algorithm}
\usepackage[noend]{algorithmic}
\usepackage[switch]{lineno}

\nolinenumbers

\newcommand{\Acronym}[1]{\ensuremath{{{\texttt{#1}}}}}
\newcommand{\Symbol}[1]{\ensuremath{\mathcal{#1}}}
\newcommand{\Function}[1]{\ensuremath{{\textsc{#1}}}}

\newcommand{\Var}[1]{\ensuremath{{{\mathrm{#1}}}}}

\newcommand{\Null}{\Acronym{null}}

\newcommand{\Pair}[1]{\ensuremath{{\langle #1 \rangle}}}
\newcommand{\World}{\Symbol{W}}
\newcommand{\Obstacles}{\Symbol{O}}
\newcommand{\Obstacle}{\Symbol{O}}
\newcommand{\Goals}{\Symbol{G}}
\newcommand{\Goal}{\Symbol{G}}
\newcommand{\Model}{\Symbol{M}}
\newcommand{\Robot}{\Symbol{R}}
\newcommand{\Shape}{\Symbol{P}}
\newcommand{\StateSpace}{\Symbol{S}}
\newcommand{\ActionSpace}{\Symbol{A}}
\newcommand{\MotionEqs}{\ensuremath{f}}
\newcommand{\Simulate}{\Function{simulate}}
\newcommand{\Traj}{\ensuremath{\zeta}}

\newcommand{\Tree}{\Symbol{T}}
\newcommand{\TreeNode}{\ensuremath{\eta}}
\newcommand{\Roadmap}{\ensuremath{\Phi}}
\newcommand{\Group}{\ensuremath{\Gamma}}
\newcommand{\CostMatrixAll}{\ensuremath{\Lambda}}
\newcommand{\CostMatrix}{\ensuremath{\lambda}}

\newcommand{\RmPoints}{\ensuremath{\Omega}}

\newcommand{\Dataset}{\ensuremath{\Xi}}
\newcommand{\DatasetEntry}{\ensuremath{\xi}}
\newcommand{\MP}{\Symbol{MP}}

\title{Combining Machine Learning and Sampling-Based Search for\\ Multi-Goal Motion Planning with Dynamics}

\author{Yuanjie Lu$^1$ \and Erion Plaku$^2$
\thanks{$^1$Department of Computer Science, George Mason University, Fairfax, VA 22030, USA.}
\thanks{The work by E. Plaku is supported by (while serving at) the National Science Foundation. Any opinion, findings, and conclusions or recommendations expressed in this material are those of the authors and do not necessarily reflect the views of the National Science Foundation.}
}

\begin{document}

\maketitle

\begin{abstract}
This paper considers multi-goal motion planning in unstructured, obstacle-rich environments where a robot is required to reach multiple regions while avoiding collisions. The planned motions must also satisfy the differential constraints imposed by the robot dynamics. To find solutions efficiently, this paper leverages machine learning, Traveling Salesman Problem (TSP), and sampling-based motion planning. The approach expands a motion tree by adding collision-free and dynamically-feasible trajectories as branches. A TSP solver is used to compute a tour for each node to determine the order in which to reach the remaining goals by utilizing a cost matrix. An important aspect of the approach is that it leverages machine learning to construct the cost matrix by combining runtime and distance predictions to single-goal motion-planning problems. During the motion-tree expansion, priority is given to nodes associated with low-cost tours. 
Experiments with a vehicle model operating in obstacle-rich environments demonstrate the computational efficiency and scalability of the approach.
\end{abstract}

\section{Introduction}
\label{sec:Intro}
Robots assisting in exploration, inspection, surveillance, warehouses, transportation, search-and-rescue, and other areas often have to reach multiple locations  to carry out assigned tasks.  These settings give rise to challenging multi-goal motion-planning problems, where the robot must quickly plan its motions to reach the goal locations while avoiding collisions. The environments are often unstructured and contain numerous obstacles, requiring the robot to navigate around them and pass through several narrow passages. These geometric problems are made harder since the planned motions must also satisfy the physical constraints imposed by the robot dynamics, which restrict the directions of motion, velocity, turning radius, and acceleration among others.

The robot also has to reason at a high level to determine appropriate orders to reach the goals, bearing similarities with Traveling Salesman Problem (TSP) formulations. In multi-goal motion planning, however, the cost, distance, or trajectory of moving from one location to another is not known beforehand. Using straight-line paths and Euclidean distances as a proxy is generally not feasible since they do not account for the obstacles or the robot dynamics. In this way, multi-goal motion planning gives rise to numerous single-goal motion-planning problems. 

Another challenge is that high-level reasoning and motion planning cannot be decoupled. Computing an order in which to reach the goals and then running a motion planner from one goal to the next in succession is unlikely to succeed since the goal ordering has not necessarily accounted for the obstacles and the robot dynamics. Thus, the spaces of feasible motions and goal orderings must be simultaneously explored.

\begin{figure}
\centering
\includegraphics[width=0.9\columnwidth]{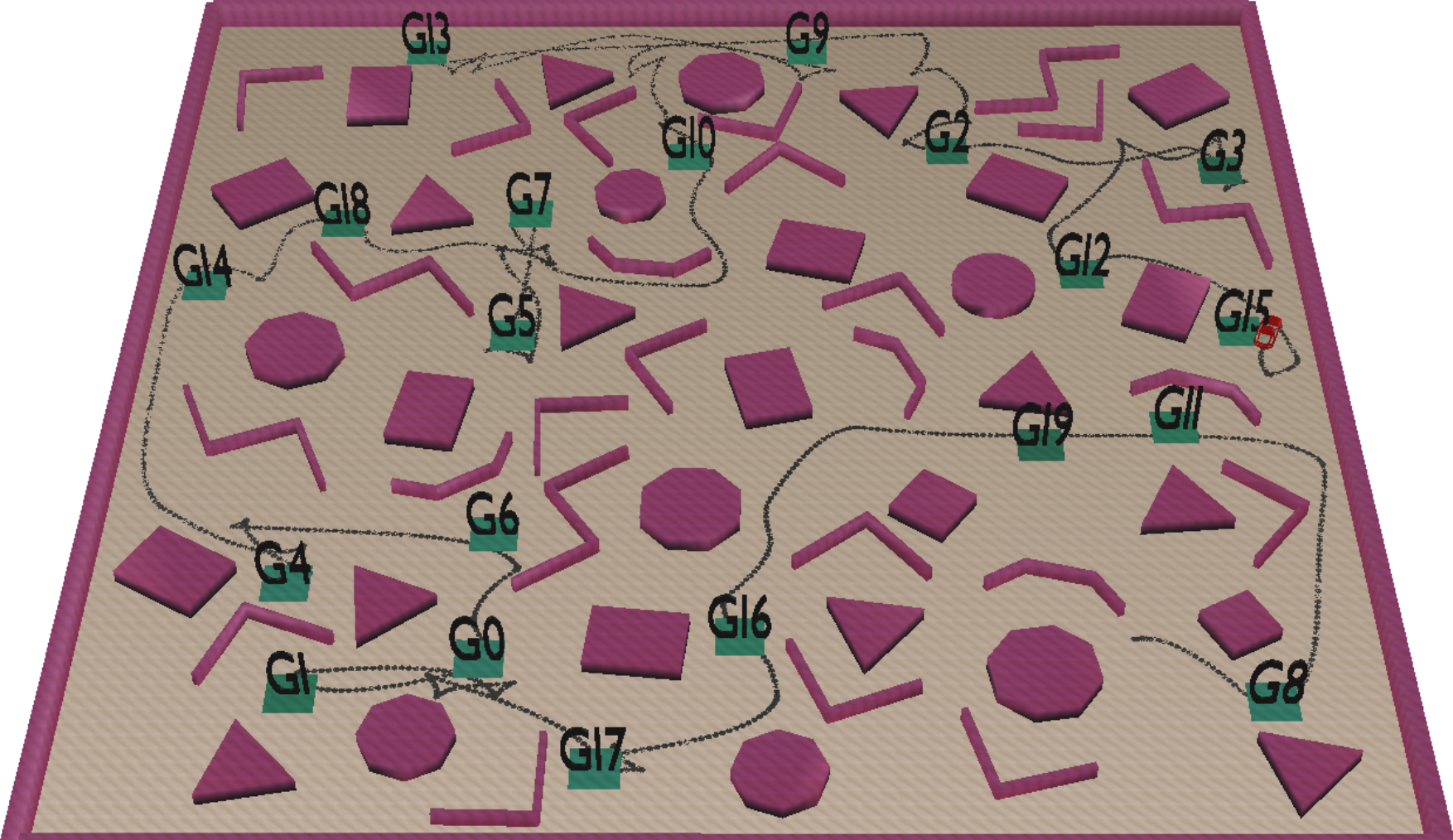}
\caption{A multi-goal motion-planning problem, where the robot has to reach each of the goals $G_0, \ldots, G_{19}$ while avoiding collisions with the obstacles. Videos of solutions obtained by our approach can be found as part of the supplementary material or by visiting the anonymous link at https://tinyurl.com/bdtsvbeu}
\label{fig:Main}
\end{figure}

To address these challenges, we develop a coupled approach that leverages sampling-based motion planning, high-level reasoning, and machine learning. We build upon related work by 
\cite{plaku2018multi}, which combined sampling-based motion planning with high-level reasoning. To account for the obstacles and the robot dynamics, 
sampling-based motion planning is leveraged to explore the space of feasible motions by expanding a motion tree whose branches correspond to collision-free and dynamically-feasible trajectories. Each node in the motion tree keeps track of the goals that have been reached along the trajectory starting from the root of the tree. The high-level reasoner is based on a TSP solver which computes a tour for each node to determine the order in which to reach the remaining goals. During the motion tree expansion, priority is given to nodes associated with low-cost tours.

In \cite{plaku2018multi}, the TSP solver relied on cost matrices based on shortest-path distances according to a roadmap graph, constructed by sampling and connecting neighboring points, similar to the Probabilistc RoadMap (PRM) method \cite{PRM}.  The roadmap provides an improvement over Euclidean distances since it takes into account the obstacles. It still falls short, however, since it does not account for the robot dynamics.

In this work, we propose to leverage machine learning to compute the cost matrices to account for obstacles, robot dynamics, and even the expected difficulty of solving the corresponding single-goal motion-planning problems. First, we train a model to predict the length of the solution trajectory for single-goal motion-planning problems by running a motion planner on thousands of randomly generated instances. The motivation is that the predictions from the trained model reflect the solutions obtained by motion planning, taking into account not only the obstacles but also the robot dynamics. Second, we train another model to predict the runtime the motion planner would take in solving single-goal motion-planning problems. The predicted runtime stands as a proxy for the problem difficulty. We combine the predicted solution distance and motion-planning runtime as a cost. In this way, the TSP solver associates a low-cost tour with each node in the motion tree to the remaining unreached goals  that reflects both the expected solution distance and the problem difficulty, taking into account the obstacles and the robot dynamics.

This work also enhances the motion-tree expansion. When attempting to expand a node in the motion tree toward the next goal in the tour, several states are sampled near the selected node, and priority is given to those states that are associated with low-cost predictions. This allows the motion tree to more effectively explore low-cost areas,  considerably reducing the motion-planning runtime.

Experiments are conducted in simulation using a vehicle model operating in unstructured, obstacle-rich environments with an increasing number of goals. Results demonstrate the computational efficiency and scalability of the approach.

\section{Related Work}

Related work has considered different aspects of the multi-goal motion-planning problem. At a high level, there has been extensive work on TSP \cite{gutin2006traveling,applegate2011traveling,helsgaun2017extension,cheikhrouhou2021comprehensive,bogyrbayeva2023deep}.



When considering the obstacles but not the robot dynamics, multi-goal problems are often solved by first constructing a roadmap based on PRM \cite{PRM} and then using a TSP solver over a cost matrix based on shortest-path distances over the roadmap to find a tour to visit the goals \cite{danner2000randomized,saha2006planning,englot2012sampling}. Ant colony optimization has also been used \cite{englot2011}.

TSP solvers have also been combined with controllers for Dubins vehicles \cite{jang2017optimal,janovs2022randomized}.

When considering the obstacles and robot dynamics, sampling-based motion planning has often been used. For multi-goal problems, sampling-based motion planning is combined with TSP solvers \cite{edelkamp2014multi,rashidian2014motion,edelkamp2018integrating,plaku2018multi,das2024motion}.

In recent years, machine learning has also been used to enhance single-goal motion planning \cite{arslan2015machine,li2018neural,ichter2020learned,wang2020neural,xiao2022motion,buiimproving,lu2023leveraging,lu2024multi}. We leverage from the work in \cite{buiimproving} the idea of training a model to predict the runtime of a motion planner on single-goal problems. We utilize the trained model to facilitate the construction of cost matrices for multi-goal motion planning and combine it with another model that can predict solution distances. 

Our work significantly improves the runtime efficiency in solving multi-goal motion planning problems, taking into account the obstacles and the robot dynamics. The runtime efficiency is due to the combination of high-level reasoning, sampling-based motion planning, and machine learning.

\section{Problem Formulation}
\label{sec:Problem}

\begin{figure*}
\centering
\includegraphics[width=0.89\textwidth]{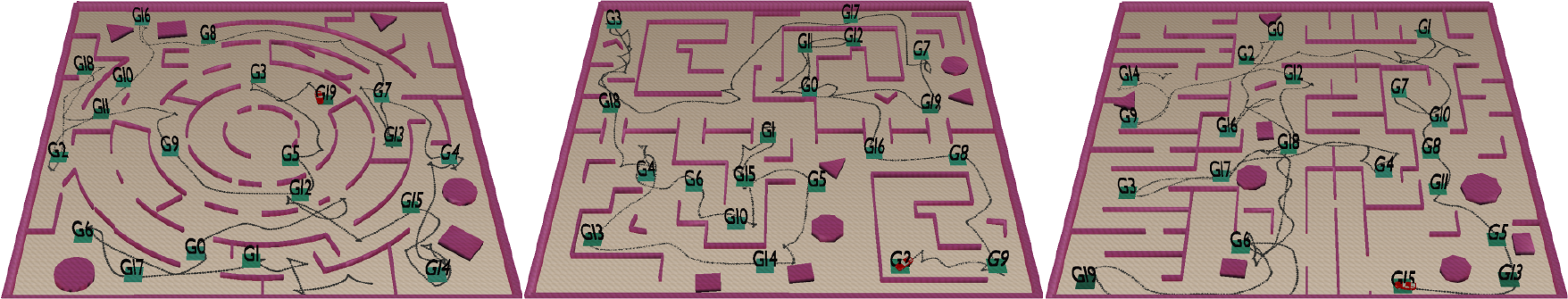}
\includegraphics[width=0.1\textwidth]{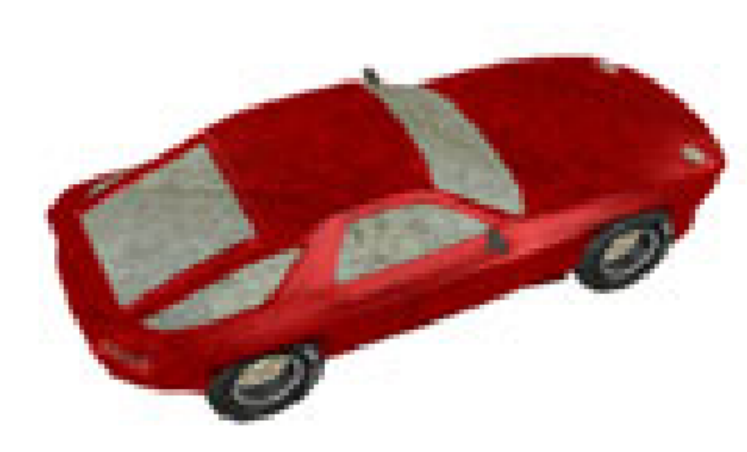}
\caption{The other scenes and the vehicle model used in the experiments (scene 1 shown in Fig.~\ref{fig:Main}).}
\label{fig:Scenes}
\end{figure*}

The robot operates in a world $\World$ that contains a set of obstacles $\Obstacles=\{\Obstacle_1, \ldots, \Obstacle_m\}$ and a set of goal regions $\Goals=\{\Goal_1, \ldots, \Goal_n\}$, as shown in Figs.~\ref{fig:Main} and \ref{fig:Scenes}. The robot model is defined as a tuple $\Robot=\Pair{\Shape, \StateSpace, \ActionSpace, \MotionEqs}$ in terms of its geometric shape $\Shape$, state space $\StateSpace$, action space $\ActionSpace$, and dynamics $\MotionEqs$ expressed as a set of differential equations of the form
\begin{equation}
\dot{s} = f(s, a), s \in \StateSpace, a \in \ActionSpace.
\end{equation}
As an example, the state for the vehicle model  (shown in Fig.~\ref{fig:Scenes}) used in the experiments is defined as $s = (x, y, \theta, \psi, v)$ in terms of the position $(x, y)$, orientation $\theta$, steering angle $\psi $, and speed $v$. The vehicle is controlled by setting the acceleration $a_\Var{acc}$ and the steering turning rate $a_\omega$. The motion equations are defined as 
\begin{equation}
\dot{x} = v\cos{(\theta)}\cos{(\psi)}, \dot{y} = v\sin{(\theta)}\cos{(\psi)}, 
\end{equation}
\begin{equation}
\dot{\theta} = v\sin{(\psi)} / L, \dot{v} = a_{acc}, \dot{\psi} = a_w, 
\end{equation}
where $L$ is the distance from the back to the front wheels.

A state $s \in \StateSpace$ is considered to have reached a goal $\Goal_i$ if the position defined by $s$ is inside $\Goal_i$.
A state $s \in \StateSpace$ is in collision if the robot overlaps with an obstacle when placed according to the position  and orientation defined by $s$.

Applying a control action $a \in \ActionSpace$ to a state $s \in \StateSpace$ yields a new state $s_\Var{new}$, which is obtained by numerically integrating the motion equations $\MotionEqs$ for one time step $dt$, i.e.,
\begin{equation}
s_{new} \gets \Simulate(s, a, f, dt).
\end{equation}
A dynamically-feasible trajectory $\Traj: \{0, \ldots, \ell\} \rightarrow \StateSpace$ is obtained by applying a sequence of actions $\Pair{a_0, \ldots, a_{\ell-1}}$ in succession, where $\Traj_0 \leftarrow s$ and $\forall j \in \{1, \ldots, \ell\}$:
\begin{equation}
\Traj_{j} \gets \Simulate(\Traj_{j-1}, a_{j-1}, f, dt).
\end{equation}
 The multi-goal motion-planning problem is defined as follows: Given a world $\World$ containing obstacles  $\Obstacles=\{\Obstacle_1, \ldots, \Obstacle_m\}$ and goal regions $\Goals=\{\Goal_1, \ldots, \Goal_n\}$, a  robot model $\Robot=\Pair{\Shape, \StateSpace, \ActionSpace, \MotionEqs}$, and an initial state $s_\Var{init} \in \StateSpace$, compute a  collision-free and dynamically-feasible trajectory $\Traj : \{0, \ldots, \ell\} \rightarrow \StateSpace$ that starts at $s_\Var{init}$ and reaches each goal in $\Goals$. The objective in this work is to reduce the planning time.

\section{Method}
\label{sec:Method}

The overall approach leverages sampling-based motion planning, TSP solvers, and machine learning. We first describe the various components of the approach (Sections~\ref{sec:MotionTree}-\ref{sec:TSP}) and then describe how to put them together (Section~\ref{sec:Overall}). Section~\ref{sec:Train} describes how the machine-learning models are trained on single-goal motion-planning problems.

\subsection{Machine-Learning Models}
\label{sec:Models}
The approach relies on a model $\Model_t$ trained to predict the average runtime of a single-goal motion planner $\MP$ when solving single-goal motion-planning problems, i.e., $\Model_t(p, g)$ predicts the average runtime $\MP$ would take in computing a collision-free and dynamically-feasible trajectory that starts at $p$ and reaches the goal $g$.  The runtime prediction serves as a proxy to determine the difficulty of single-goal motion-planning problems. The multi-goal approach uses the runtime predictions to steer the search toward the easier single-goal motion-planning problems so that it can improve the overall runtime efficiency. The approach also leverages another model $\Model_d$ trained to predict the distance of solution trajectories computed by $\MP$.  The runtime and solution distance predictions are combined into a cost function, i.e.,
\begin{equation}
\Function{FromToCost}(p, g) = \alpha \Model_d(p, g) + (1 - \alpha) \Model_t(p, g),
\label{eqn:FromToCost}
\end{equation}
where $0 \leq \alpha \leq 1$ provides a parameter to fine-tune the importance of each component. Section~\ref{sec:Train} provides details on how the models $\Model_\Var{time}$ and $\Model_\Var{dist}$ are trained.

\subsection{Motion Tree}
\label{sec:MotionTree}

To account for the obstacles and the robot dynamics, a motion tree $\Tree$ is rooted at the initial state $s_\Var{init}$ and is incrementally expanded in $\StateSpace$. Each node $\TreeNode \in \Tree$ is associated with the fields $\Pair{\Var{parent}, s, a, \Var{goals}}$. The parent node is indicated by $\TreeNode.\Var{parent}$. The state $\TreeNode.s \in \StateSpace$ is obtained by applying the control action $\TreeNode.a$ to the state of the parent node for one time step, i.e., $\TreeNode.s \gets \Simulate(\TreeNode.\Var{parent}.s, \TreeNode.a, \MotionEqs, dt)$. By construction, $\TreeNode$ is added to $\Tree$ only if $\TreeNode.s$ is not in collision.

Let $\Traj_\Tree(v)$ denote the trajectory obtained as the sequence of states from the root of $\Tree$ to $\TreeNode$. The node $\TreeNode$ also keeps track of the goals reached by $\Traj_\Tree(v)$, denoted by $\TreeNode.\Var{goals}$. A solution is found when $\Traj_\Tree(v)$ reaches all the goals, i.e., $\TreeNode.\Var{goals} = \Goals$.

The motion-tree expansion is guided by two critical decisions at each iteration: (i) which node $\TreeNode$ to select from $\Tree$, and (ii) how to expand $\Tree$ from $\TreeNode$? For the second question, the objective is to expand $\Tree$ to reach the remaining goals, i.e., $\Goals\setminus \TreeNode.\Var{goals}$. We can rely on a TSP solver to determine an appropriate order for the remaining goals. This requires defining the cost from $\TreeNode.s$ to each remaining goal and the costs for each pair of remaining goals to construct a cost matrix that the TSP solver can use. This is where machine learning comes into play, as we can train models to predict these costs, noting that each matrix entry corresponds to a single-goal motion-planning problem. For the first question, we can use the tour cost to give priority to nodes in $\Tree$ associated with low-cost tours when deciding from which node to expand $\Tree$.

\subsection{Motion-Tree Partition into Groups}
As presented, the approach would be too slow since $\Tree$ often has thousands of nodes, so solving a TSP for each node would be impractical. To address this, we group the nodes in $\Tree$ based on their vicinity and the goals that they have reached. Specifically, we generate a number (several hundred in the experiments) of collision-free points $\Roadmap$. We augment the fields associated with each node in $\Tree$ to also keep track of the nearest point in $\Roadmap$, i.e., $\TreeNode.p \gets \Function{nearest}(\Roadmap, \Function{position}(\TreeNode.s))$.

A group $\Group_\Pair{p, \Var{goals}}$ with $p \in \Roadmap$ and $\Var{goals} \subseteq \Goals$ is defined to include all the nodes in $\Tree$ that have reached the indicated goals and have $p$ as their nearest point, i.e.,
\begin{eqnarray}
\Group_\Pair{p, \Var{goals}}.\Var{nodes} = \{ \TreeNode : \TreeNode \in \Tree \wedge \TreeNode.\Var{goals} = \Var{goals} \wedge  \TreeNode.p = p\}.
\end{eqnarray}
This induces a partition of $\Tree$  into a set of groups $\Group$. Specifically, when a node $\TreeNode$ is added to $\Tree$, $\Group$ is searched to determine whether it contains the group $\Group_\Pair{\TreeNode.p, \TreeNode.\Var{goals}}$. If so, $\TreeNode$ is added to the group. Otherwise, the group is created, $\TreeNode$ is added to the group, and the group is added to $\Group$.

\subsection{TSP Tours for the Motion-Tree Groups}
\label{sec:TSP}
When a group $\Group_\Pair{p, \Var{goals}}$ is first created, we also invoke a TSP solver to determine a tour in which to visit the remaining goals $\Goals\setminus \Var{goals}$ starting from $p$, i.e.,
\begin{equation}
 \Group_\Pair{p, \Var{goals}}.\Var{tour} \gets \Function{TSPsolver}(\Function{CostMatrix}(p, \Goals\setminus\Var{goals}))
\end{equation}
An entry $\CostMatrix_{i,j}$ in the cost matrix corresponds to a single-goal motion-planning problem (from $p$ to a goal or from a goal to another goal). We can use the machine-learning models to predict the cost of each entry, i.e., 
\begin{equation}
\CostMatrix_{i, j} \gets \Function{FromToCost}(i, j).
\end{equation}
In this way, the TSP solver will compute tours that take into account both the difficulty in solving the corresponding single-goal motion-planning problems and the expected length of the solution trajectories. We use LKH-3.0 \cite{helsgaun2017extension} as the TSP solver. Note that the tours are open since there is no requirement to return to the start node.

\subsection{Putting it all Together: Multi-Goal Motion Planning Augmented with Machine Learning}
\label{sec:Overall}
Pseudocode for the overall approach is shown in Alg.~\ref{alg:Main}. The motion tree $\Tree$ is rooted at the initial state $s_\Var{init}$ (Alg.~\ref{alg:Main}:1). A number of collision-free samples $\Roadmap$ (including the goal centers) are then generated,  which are used to induce the partition of $\Tree$ into groups (Alg.~\ref{alg:Main}:2-3). The partition initially contains only one group, which has only the initial node (Alg.~\ref{alg:Main}:4).

 Each iteration starts by first selecting the group with the maximum weight from the motion-tree partition $\Group$ (Alg.~\ref{alg:Main}:6). The weight of a group $\Group_{p, \Var{goals}}$ is defined as 
 \begin{equation}
\Group_{p, \Var{goals}}.w = {\beta^{\Group_{p, \Var{goals}}.\Var{nrSel}}\,\gamma^{|\Var{goals}|}} / { \Function{cost}(\Group_{p, \Var{goals}}.\Var{tour}) },
 \label{eqn:GroupWeight}
 \end{equation}
 where $0 < \beta < 1$, $\Group_{p, \Var{goals}}.\Var{nrSel}$ denotes how many times the group has been selected, and $\gamma > 1$. This gives priority to groups that have reached many goals and have low-cost tours to the remaining goals since they are most likely to lead to the solution quickly. The number of selections provides a penalty  to counter the greediness of the tour costs to avoid becoming stuck when expansions from  $\Group_{p, \Var{goals}}$ are infeasible due to constraints imposed by obstacles and robot dynamics.

After selecting a group $\Group_{p, \Var{goals}}$, a node $\TreeNode$ is then selected at random from $\Group_{p, \Var{goals}}.\Var{nodes}$ (Alg.~\ref{alg:Main}:7). The objective is to expand $\TreeNode$ toward the first goal $g$ in the tour associated with $\Group_{p, \Var{goals}}$ (Alg.~\ref{alg:Main}:8, see also Section~\ref{sec:TSP} on how tours are computed using a TSP solver and the costs predicted by the machine-learning models). To make progress from $\TreeNode$ toward $g$, a target $p_\Var{target}$ is selected near $\TreeNode$ in a direction that is estimated to reduce the cost (Alg.~\ref{alg:Main}:9).
As shown in Alg.~\ref{alg:Main}(b), several points are sampled near $\TreeNode$. For each sample  $p$, the overall cost is estimated by using the machine-learning models to predict the cost from $\TreeNode$ to $p$ and from $p$ to $g$. The sample with the minimum cost is then selected as the target.

 $\Function{extend}(\Tree, \TreeNode, p_\Var{target})$ extends $\Tree$ from $\TreeNode.s$ toward  $p_\Var{target}$ (Alg.~\ref{alg:Main}:9 and Alg.~\ref{alg:Main}(c)). Specifically, proceeding iteratively, a PID controller selects appropriate control actions at each iteration to steer the robot from $\TreeNode.s$ toward $p_\Var{target}$, e.g., by turning the wheels toward $p_\Var{target}$. Each new state $s_\Var{new}$ (obtained by applying the selected control to the parent state for one time step) is checked for collisions. If not in collision, it is added as a new node $\TreeNode_\Var{new}$ to $\Tree$. The new node inherits the goals from its parent, and if $s_\Var{new}$ reaches a new goal, it is added to $\TreeNode_\Var{new}.\Var{goals}$. The nearest sample from $\Roadmap$ is also computed since this information is used for the motion-tree partition. If $s_\Var{new}$ is in collision, the extension toward $p_\Var{target}$ terminates. The extension also terminates when $s_\Var{new}$ is close to $p_\Var{target}$ or a maximum number of iteration steps is reached.

The motion-tree partition $\Group$ is updated to account for the new nodes added by $\Function{extend}$ (Alg.~\ref{alg:Main}:11-13 and Alg.~\ref{alg:Main}(a)). As described in Section~\ref{sec:TSP}, the TSP solver is invoked to compute a tour each time a new group is added to $\Group$ by utilizing the machine-learning models to build the cost matrix.

\begin{algorithm}[!t]
\caption{MultiGoalMP-ML}
\label{alg:Main}
\textbf{Input}: world $\mathcal{W}$; obstacles $\mathcal{O}$; goals $\mathcal{G}$; robot model $\mathcal{R}$; initial state $s_\Var{init}$; $\Function{FromToCost}$ to estimate cost (combined runtime and solution distance) of single-goal motion planning;  $\Function{TSPsolver}$ to solve TSP problems; runtime limit $t_\Var{max}$ \\ 
\textbf{Output}: collision-free and dynamically-feasible trajectory that starts at $s_\Var{init}$ and reaches each goal in $\Goals$, or $\Null$ if no solution is found within $t_\Var{max}$ runtime
\vspace*{1mm}
\hrule
\begin{algorithmic}[1]
\STATE $\Tree \gets \Function{CreateMotionTree}(s_\Var{init})$
\STATE $\Roadmap \gets \Function{GenerateCollisionFreeSamples}(\World, \Obstacles, \Goals)$
\STATE \textbf{for} $g \in \Goals$ \textbf{do} $\Function{add}(\Roadmap, \Function{center}(g))$
\STATE $\Group \gets \emptyset$; $\Function{UpdateGroups}(\Group, \TreeNode_\Var{init}, \CostMatrixAll)$
\WHILE{$\Function{time} < t_\Var{max}$}
\STATE $\Group_{p, \Var{goals}} \gets \Function{SelectGroup}(\Group)$
\STATE $\TreeNode \gets \Function{SelectNode}(\Group_{p, \Var{goals}}.\Var{nodes})$
\STATE $g \gets$ first goal in $\Group_{p, \Var{goals}}.\Var{tour}$
\STATE $p_\Var{target} \gets \Function{SelectTarget}(\TreeNode, g)$
\STATE $\Function{extend}(\Tree, \TreeNode, p_\Var{target})$
\FOR{each new node $\TreeNode_\Var{new}$ added by $\Function{extend}$}
\STATE \textbf{if} $\TreeNode_\Var{new}.\Var{goals} = \Goals$ \textbf{return} $\Traj_\Tree(\TreeNode_\Var{new})$
\STATE $\Function{UpdateGroups}(\Group, \TreeNode_\Var{new})$
\ENDFOR
\ENDWHILE

\end{algorithmic}

\hrule
\vspace*{1mm}
(a) $\Function{UpdateGroups}(\Group, \TreeNode_\Var{new})$
\begin{algorithmic}[1]
\STATE $\Var{key} \gets \Pair{\TreeNode_\Var{new}.p, \TreeNode_\Var{new}.\Var{goals}}$
\STATE $\Group_\Var{key} \gets \Function{find}(\Group, \Var{key})$

\IF{$\Group_\Var{key} = \Null$}
\STATE $\Group_\Var{key} \gets \Function{NewGroup}(\Var{key})$; $\Function{AddGroup}(\Group, \Group_\Var{key})$
\STATE $V \gets \emptyset$; $\Function{add}(V, \TreeNode_\Var{new}.p)$
\STATE \textbf{for} $g \in \Goals\setminus \TreeNode_\Var{new}.\Var{goals}$ \textbf{do} $\Function{add}(V, g)$
\STATE \textbf{for} {$(i, j) \in V \times V$} \textbf{do} $\CostMatrix_{i, j} \gets \Function{FromToCost}(i, j)$
\STATE $\Group_\Var{key}.\Var{tour} \gets \Function{TSPsolver}(\CostMatrix)$
\ENDIF

\STATE $\Function{AddNode}(\Group_\Var{key}, \TreeNode_\Var{new})$
\end{algorithmic} 

\hrule
\vspace*{1mm}
(b) $\Function{SelectTarget}(\TreeNode, g)$
\begin{algorithmic}[1]
\STATE $c_\Var{min} \gets \infty$; $p_\Var{target} \gets \Null$
\FOR{several times}
\STATE $p \gets \Function{GenerateSampleInVicinity}(\TreeNode)$
\STATE $c \gets \Function{FromToCost}(\TreeNode, p) + \Function{FromToCost}(p, g)$
\STATE \textbf{if} $c < c_\Var{min}$ \textbf{then} $\{ c_\Var{min} \gets c; p_\Var{target} \gets p\}$
\ENDFOR
\STATE \textbf{return} $p_\Var{target}$
\end{algorithmic}

\hrule
\vspace*{1mm}
(c) $\Function{extend}(\Tree, \TreeNode, p_\Var{target})$
\begin{algorithmic}[1]
    \FOR{several steps}
    \STATE $a \gets \Function{controller}(\TreeNode.s, p_\Var{target})$
    \STATE $s_\Var{new} \gets \Simulate(\TreeNode.s, a, \MotionEqs, dt)$
    \STATE \textbf{if} $\Function{collision}(s_\Var{new})$ \textbf{then break}
    \STATE $\TreeNode_\Var{new} \gets \Function{NewNode}()$; $\Function{add}(\Tree, \TreeNode_\Var{new})$
    \STATE $\TreeNode_\Var{new}.\Pair{\Var{parent}, s, a} \gets   \Pair{\TreeNode, s_\Var{new}, a}$
    \STATE $\TreeNode_\Var{new}.\Var{goals} \gets   \TreeNode.\Var{goals} \cup \{\Function{GoalReached}(\Goals, s_\Var{new})\}$
    \STATE $\TreeNode_\Var{new}.p \gets   \Function{nearest}(\Roadmap, \Function{position}(s_\Var{new}))$
    \STATE \textbf{if} $\Function{reached}(s_\Var{new}, p_\Var{target})$ \textbf{then break} 
    \STATE $\TreeNode\gets \TreeNode_\Var{new}$
    \ENDFOR
\end{algorithmic} 
\end{algorithm}

\subsection{Training on Single-Goal Motion Planning}
\label{sec:Train}

As described, our approach relies on models trained to predict the runtime and solution distance of a motion planner $\MP$ on single-goal motion-planning problems. The training is conducted offline and relies on generated datasets. Given a world $\World$ with obstacles $\Obstacles$, a robot model $\Robot$, and a single-goal motion planner $\MP$, we create a training dataset $\Dataset_{\World, \Obstacles, \Robot, \MP}$. When the context is clear, $\Dataset_{\World, \Obstacles, \Robot, \MP}$ is referred to as $\Dataset$.

An instance $\DatasetEntry \in \Dataset$ is a tuple $\Pair{p, g, t, d}$ where $p$ is the start position, $g$ is the goal position, and $t$ and $d$ denote the average runtime and solution distance as obtained by  $\MP$. The objective of $\MP$ is to compute a collision-free and dynamically-feasible trajectory that starts at $p$ and reaches a small square centered at $g$. The average runtime and solution distance are obtained by running $\MP$ twenty times for each pair $\Pair{p, g}$. Each run continues until a solution is found or a time limit (set to $10$s in the experiments) is reached. To avoid the influence of outliers, 20\% of the runs with the lowest and highest runtimes are removed when computing the averages.

For the single-goal motion planner $\MP$, we used a modified version of Alg.~\ref{alg:Main}. Specifically, when invoked as $\MP(p, g)$, the position of the initial state is set to $p$ (the other state components are set to $0$) and the set of goals $\Goal$ consisted only of $g$. The machine-learning component is removed. To estimate the from-to cost, first the collision-free samples $\Roadmap$ were connected to form a roadmap as in PRM. Then, $\Function{FromToCost}(a, b)$ is computed as the shortest-path distance in the roadmap.
We also show results (Section~\ref{sec:ExpRes}) when using the single-goal version of the multi-goal motion planner DROMOS \cite{plaku2018multi}  as the $\MP$.

The work in \cite{buiimproving} only focused on predicting the runtime, but not on the solution distances. Another distinction is how the single-goal motion-planning problems are generated. The work in \cite{buiimproving} generated each $p$ and $g$ at random (repeatedly sampling until no collision). For our work, we first sampled a set $\RmPoints$ of collision-free points ($|\RmPoints|=500$ for the experiments). We then ran $\MP$ twenty times on each pair in $\RmPoints \times \RmPoints$ (so $|\Dataset|=25,000$) to obtain the average runtime and solution distance. This provided a more robust dataset since each point was paired with many others.

\begin{figure}[!t]
\centering
\includegraphics[width=\columnwidth]{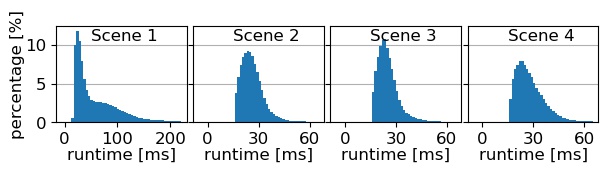}\\ [-1.5mm]
\includegraphics[width=\columnwidth]{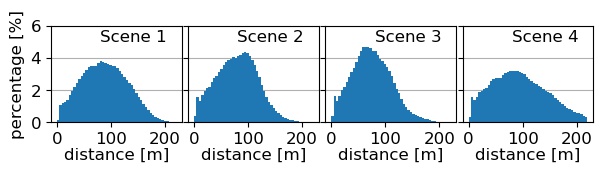}
\caption{Runtime and solution distance distributions for single-goal motion-planning problems.}
\label{fig:DataDistribution}
\end{figure}

Fig.~\ref{fig:DataDistribution} shows the runtime and solution distance distribution of each of the four datasets (one for each scene). The Gaussian distribution is beneficial for exploratory data analysis. As expected, a large number of the instances are easy to solve. 

We compared three machine-learning methods: Artifical Neural Network (ANN), XGBoost \cite{chen2016xgboost}, and LightGBM \cite{ke2017lightgbm}. The ANN model was the same as in \cite{buiimproving}. For XGBoost, the best results were obtained by using the default parameters except for the maximum number of leaves (set to $16$) and the maximum depth (set to $12$). To prevent high regression errors and slow inference time, the learning rate of the model was set to $0.5$ and the training epoch to $48$. LightGBM has the same parameters and the same inference model size as XGBoost. 

\begin{figure}[!t]
\centering
\includegraphics[width=\columnwidth]{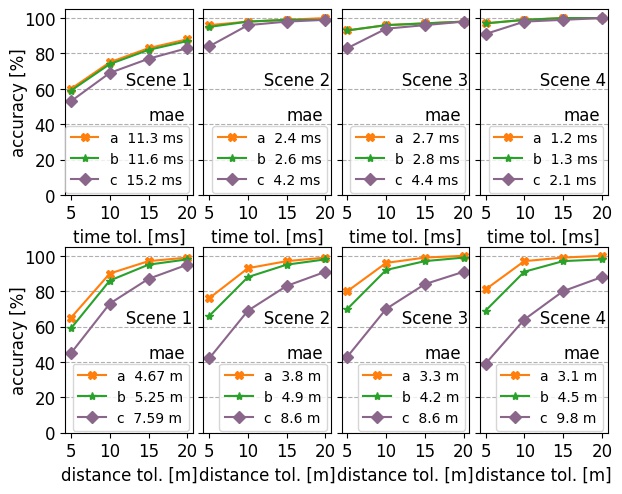}\\
\caption{Prediction accuracy of (a) XGBoost, (b) LightGBM, and (c) ANN on the single-goal motion-planning instances. The accuracy is computed as the percentage of test instances whose predicted value falls within the indicated tolerance (80\%-20\% split of the data for training and testing).}
\label{fig:Accuracy}
\end{figure}

Fig.~\ref{fig:Accuracy} shows that XGBoost achieves a higher accuracy for both the runtime and solution distance predictions. As a result, our overall approach used the XGBoost models.

\begin{figure*}[!t]
  \centering  \includegraphics[width=\textwidth]{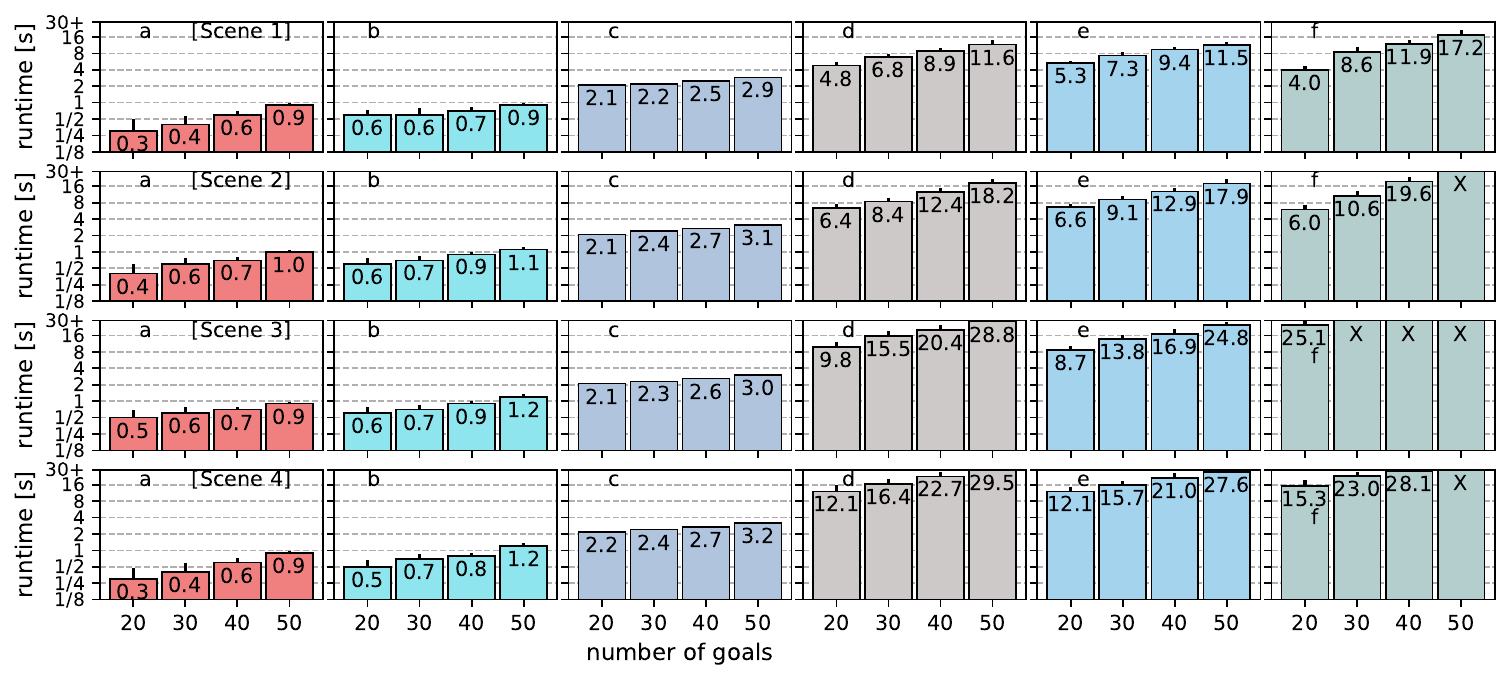}
   \caption{Runtime results when varying the number of goals: (a) our approach, MultiGoalMP-ML (Alg.~\ref{alg:Main}); (b) MultiGoalMP-ML[DROMOS-SingleGoal], which refers to our approach when using the single-goal version of DROMOS to generate the training datasets for the machine-learning components of our approach; (c) DROMOS; (d) MultiGoalMP-RM; (e) MultiGoalMP-ED; and
(f) DROMOS-Random. The runtime includes everything from preprocessing to reporting that a solution is found. A time limit of $30$s was imposed for each run. Entries marked with $X$ indicate failure (reported runtime reached the limit).}
  \label{fig:ResRuntimeGoals}
\end{figure*}

\begin{figure*}[!t]
  \centering
  \includegraphics[width=2.0\columnwidth]{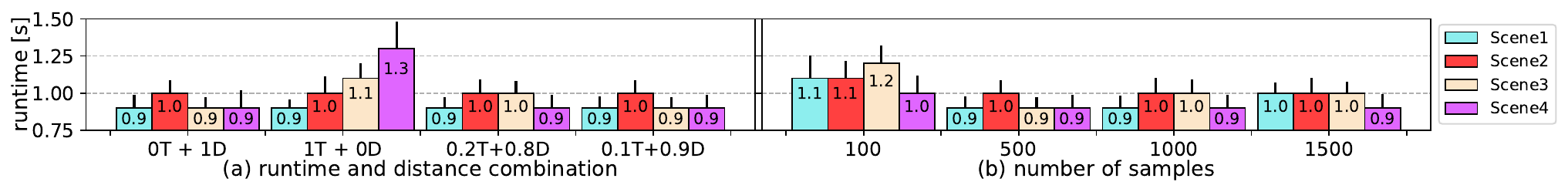}\\[-1mm]
  \includegraphics[width=2.0\columnwidth]{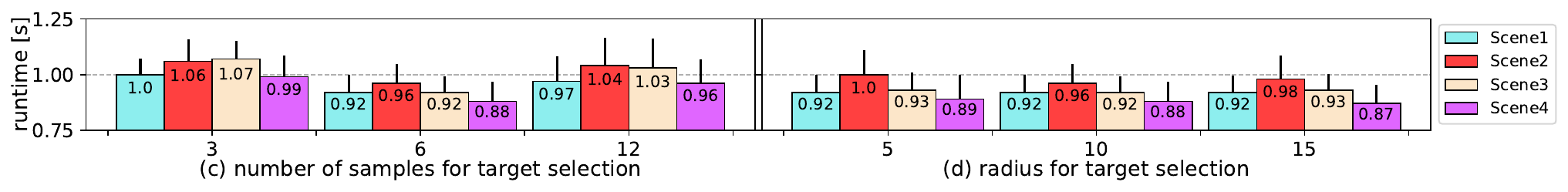}
  \caption{Impact of various parameters on MultiGoalMP-ML (Alg.~\ref{alg:Main}): (a) coefficient $\alpha$ when combining the runtime and solution-distance predictions for the function $\Function{FromToCost}$ (Eqn.~\ref{eqn:FromToCost}); (b) number of collision-free samples used for the motion-tree partition into groups (Alg.~\ref{alg:Main}:2); (c) number of samples used for the target selection (Alg.~\ref{alg:Main}(c)); and (d) radius for the target selection (Alg.~\ref{alg:Main}(c)).}
  \label{fig:ResParams}
\end{figure*}

\section{Experiments and Results}
\label{sec:ExpRes}
Experiments are conducted using a vehicle model operating in unstructured, obstacle-rich environments. The scenes are shown in Fig.~\ref{fig:Main} and \ref{fig:Scenes}, while the robot model is described in Section~\ref{sec:Problem}. These represent challenging motion-planning problems where the robot has to carefully move around obstacles and pass through narrow passages to reach the goals. 

\subsection{Experimental Setup}

\subsubsection{Methods used for Comparisons}
Our approach, MultiGoalMP-ML (Alg.~\ref{alg:Main}), is compared to DROMOS \cite{plaku2018multi}, a state-of-the-art motion planner specifically designed for multi-goal problems. DROMOS was shown to be significantly faster than other planners. DROMOS uses a roadmap, shortest roadmap paths, and a TSP solver to guide the motion-tree expansion. As a baseline, we also used DROMOS-Random, which replaces the TSP solver with a random ordering of the remaining goals. 

As described in Section~\ref{sec:Train}, the training datasets for our approach were generated using a modified version of Alg.~\ref{alg:Main}. 
To show that our approach can be used with different single-goal motion planners for generating the training datasets, we also included experiments when using the single-goal version of DROMOS to generate the datasets. This version is referred to as MultiGoalMP-ML[DROMOS-SingleGoal].

To show the importance of the machine-learning components, we also compared our approach to modified versions of Alg.~\ref{alg:Main} that do not rely on machine learning. The first modification, referred to as MultiGoalMP-ED, is obtained by implementing $\Function{FromToCost}(a, b)$ as the Euclidean distance from $a$ to $b$. The second version, referred to as MultiGoalMP-RM, constructs a roadmap (by connecting the samples generated in Alg.~\ref{alg:Main}:2-3) and implementing $\Function{FromToCost}(a, b)$ as the distance of the shortest path from $a$ to $b$ in the roadmap.

\subsubsection{Multi-Goal Motion-Planning Problem Instances}
Experiments are conducted with $20, 30, 40,$ and $50$ goals.  For a given scene and number of goals, we generate 200 problem instances by placing the goals at random collision-free locations. Each method is run on each of the instances. An upper bound of $30$s is set for each run. When calculating the performance statistics, we remove the instances within 20\% of the minimum runtime and within 20\% of the maximum runtime to avoid the influence of outliers. The runtime includes everything from reading the input files to reporting that a solution is found. The solution length is measured as the distance traveled by the robot along the solution trajectory. 

\subsubsection{Computing Resources}
The experiments were run on an AMD Ryzen 9 5900X (CPU: 3.7 GHz) using Ubuntu 20.04. The motion-planning code was written in C++ and compiled using g++-9.4.0. The code for training the models was written in Python 3.8 using the 1.7.2 version of XGBoost. The learned models were imported to C++ via the C API of treelite~\cite{cho2018treelite}.

\subsection{Performance Evaluation}

\subsubsection{Runtime Results when Increasing the Number of Goals}
Fig.~\ref{fig:ResRuntimeGoals} shows the runtime results when varying the number of goals from $20$ to $50$.  The results indicate that MultiGoalMP-ML (Alg.~\ref{alg:Main}) is considerably faster than all the other methods. This is due to the effective interplay between sampling-based motion planning and machine learning. The predictions make it possible to guide the motion-tree expansion along low-cost tours which are associated with easier-to-solve single-goal motion-planning problems. Our approach, MultiGoalMP-ML, also works well when using the single-goal version of DROMOS to generate the training datasets used by the machine-learning components.

DROMOS is considerably slower than MultiGoalMP-ML. DROMOS relies on tours based on shortest-path distances over the roadmap, which could be difficult to follow due to the constraints imposed by the obstacles and robot dynamics. This is also the case for MultiGoalMP-RM, the variant of our approach that removes the machine-learning component and uses shortest-paths over the roadmap as the from-to-cost estimate. Similar observation also holds for MultiGoalMP-ED, which uses the Euclidean distance as the from-to-cost estimates. As expected, DROMOS-Random has the worst performance since it relies on tours based on random permutations of the remaining goals.

\subsubsection{Relative Performance on Runtime and Solution Distance}
Table~\ref{table:ResTradeoff} shows the tradeoff between runtime and solution distances. As expected, DROMOS is able to find shorter solutions since it is guided by tours based on shortest-path distances over the roadmap. The relative increase for the solution distances for MultiGoalMP-ML when compared to DROMOS ranges from $0.35$ to $0.66$, while the relative increase of the runtime of DROMOS compared to MultiGoalMP-ML ranges from $2.1$ to $6.3$. Thus, MultiGoalMP-ML offers significant improvements in runtime with modest increases in solution length.

\subsubsection{Impact of Various Parameters}
MultiGoalMP-ML has several parameters that can be used to fine-tune its performance. Unless otherwise indicated, MultiGoalMP-ML is run with the following parameter values: $\alpha = 0.9$ (Eqn.~\ref{eqn:FromToCost}), $\beta=0.99$ and $\gamma = 8$ (Eqn.~\ref{eqn:GroupWeight}), $|\Roadmap|=500$ (Alg.~\ref{alg:Main}:2), number of samples and radius for target selection are $6$ and $10$, respectively (Alg.~\ref{alg:Main}(b)). 

Fig.~\ref{fig:ResParams}(a) shows the results when varying the coefficient $\alpha$ to combine the runtime and solution-distance predictions for  $\Function{FromToCost}$ (Eqn.~\ref{eqn:FromToCost}). The results indicate that it is better to combine the predictions than to just use one or the other. Fig.~\ref{fig:ResParams}(b) shows the results for different numbers of collision-free samples (Alg.~\ref{alg:Main}:2), while  Fig.~\ref{fig:ResParams}(c-d) shows the results when varying the number of samples and the radius for the target selection (Alg.~\ref{alg:Main}(c):3). These results indicate that the approach works well for a wide range of parameter values.

\begin{table}
\begin{tabular}{c|cccc|cccc}
& \multicolumn{4}{c|}{(a) Runtime} 
& \multicolumn{4}{c}{(b) Solution Distance}\\\hline
Goals & S1 & S2 & S3 & S4 & S1 & S2 & S3 & S4\\
20 & 6.0 & 4.2 & 3.2 & 6.3 & 0.36  & 0.43 & 0.54  & 0.42 \\
30 & 4.5 & 3.0 & 2.8 & 5.0 & 0.39  & 0.44 &  0.60 & 0.46 \\
40 & 3.1 & 2.8 & 2.7 & 3.5 & 0.42  &  0.46 &  0.63 & 0.49 \\
50 & 2.2 & 2.1 & 2.3 & 2.5 & 0.46  & 0.50 & 0.66  &  0.59\\\\
\end{tabular}
\caption{(a) Relative increase in runtime by DROMOS when compared to MultiGoalMP-ML, measured as $(t_\Var{DROMOS} - t_\Var{MutliGoalMP-ML})/t_\Var{MultiGoalMP-ML}$.
(b) Relative increase in solution distance by MultiGoalMP-ML when compared to DROMOS, measured as $(d_\Var{MultiGoalMP-ML} - d_\Var{DROMOS})/d_\Var{DROMOS}$. Scenes 1-4 are denoted as S1, S2, S3, and S4.
}
\label{table:ResTradeoff}
\end{table}

\begin{figure}[!t]
\centering
\includegraphics[width=\columnwidth]{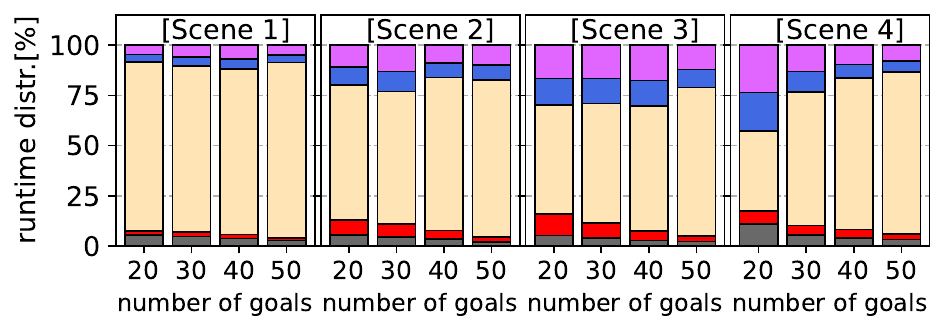}\\
\caption{Runtime distribution as the percentage of time taken by various components of MultiGoalMP-ML (Alg.~\ref{alg:Main}). From bottom to top: (a) generation of the collision-free samples (Alg.~\ref{alg:Main}:2-3); (b) machine-learning predictions, i.e., $\Function{FromToCost}$; (c) $\Function{TSPsolver}$; (d) $\Function{collision}$ and $\Function{simulate}$;  (e) other.}
\label{fig:ResRuntimeDistribution}
\end{figure}

\subsubsection{Runtime Distribution}
Fig.~\ref{fig:ResRuntimeDistribution} shows the runtime distribution for the various components of MultiGoalMP-ML (Alg.~\ref{alg:Main}). The results show the efficiency of the machine-learning predictions, which end up taking only a small percentage of the overall runtime. Most of the runtime is taken by $\Function{TSPsolver}$ since it is invoked numerous times to compute tours for each new group added to the motion-tree partition. The time dedicated to the motion-tree expansion (collision detection and simulation of robot dynamics) constitutes only a small percentage. In this way, MultiGoalMP-ML is able to shift the burden to the high level  (the TSP-guided tours and machine-learning predictions), which is considerably faster than having to explore large parts of the continuous state space. Thus, MultiGoalMP-ML focuses the exploration on the relevant parts of the state space, enabling the approach to quickly find a collision-free and dynamically-feasible trajectory that reaches all the goals.

\section{Discussion}

This paper developed an effective approach for multi-goal motion planning  in unstructured, obstacle-rich environments, taking into account the underlying robot dynamics. The effectiveness of the approach derived from coupling sampling-based motion planning with TSP solvers and machine learning. We trained models to predict the runtime and solution distance on single-goal motion-planning problems and used the trained models to guide the motion-tree expansion.

This paper opens up several avenues for research. One direction is to improve the prediction capabilities of the models so that they can also predict the solution trajectories. The challenge is that the predictions should be computationally very efficient as the trained models would be required to make many predictions when solving a multi-goal motion-planning problem. Another direction is to generalize training so that models trained in one environment can be used in other (similar) environments. Future work could also include extending the framework to multi-goal motion planning with time windows, pickups, and deliveries as it is often the case in warehouse and logistics operations.

\bibliographystyle{named}
\bibliography{ijcai19}

\end{document}